\documentclass{article}
\usepackage{amssymb}
\usepackage{graphicx}

\PassOptionsToPackage{numbers,sort&compress}{natbib}
\usepackage[preprint]{corl_2025}

\usepackage[numbers,sort&compress]{natbib}
\usepackage{graphicx}
\usepackage{subfigure}
\usepackage{amsmath}
\usepackage{amssymb}
\usepackage{amsfonts}
\usepackage{bbm}
\usepackage{enumitem}
\usepackage{wrapfig}
\usepackage{booktabs}
\usepackage{capt-of}
\usepackage{hyperref}

\definecolor{orangecolor}{RGB}{255,127,0}
\hypersetup{
    colorlinks=true,
    linkcolor=blue,
    urlcolor=orangecolor
    }

\title{CogniPlan: Uncertainty-Guided Path Planning with Conditional Generative Layout Prediction}

\author{
    Yizhuo Wang$^{1}$\quad
    Haodong He$^{2}$\quad
    Jingsong Liang$^{1}$\quad
    Yuhong Cao$^{1\dagger}$\\
    \vspace{0.1cm}
    \textbf{Ritabrata Chakraborty}$^{3}$\quad
    \textbf{Guillaume Sartoretti}$^{1}$\\
    $^{1}$National University of Singapore\quad
    $^{2}$Tongji University\quad
    $^{3}$BITS Pilani
}

\begin{document}
\maketitle
\footnotetext{$^\dagger$Corresponding author. \textbf{Project page:} \url{https://yizhuo-wang.com/cogniplan/}}

\vspace{-0.5cm}
\begin{abstract}
Path planning in unknown environments is a crucial yet inherently challenging capability for mobile robots, which primarily encompasses two coupled tasks: autonomous exploration and point-goal navigation.
In both cases, the robot must perceive the environment, update its belief, and accurately estimate potential information gain on-the-fly to guide planning.
In this work, we propose CogniPlan, a novel path planning framework that leverages multiple plausible layouts predicted by a \underline{co}nditional \underline{g}e\underline{n}erative \underline{i}npainting model, mirroring how humans rely on cognitive maps during navigation.
These predictions, based on the partially observed map and a set of layout conditioning vectors, enable our planner to reason effectively under uncertainty. 
We demonstrate strong synergy between generative image-based layout prediction and graph-attention-based path planning, allowing CogniPlan to combine the scalability of graph representations with the fidelity and predictiveness of occupancy maps, yielding notable performance gains in both exploration and navigation.
We extensively evaluate CogniPlan on two datasets (hundreds of maps and realistic floor plans), consistently outperforming state-of-the-art planners.
We further deploy it in a high-fidelity simulator and on hardware, showcasing its high-quality path planning and real-world applicability.

\end{abstract}


\section{Introduction}

\begin{wrapfigure}{r}{0.56\textwidth}
    \centering
    \vspace{-0.6cm}
    \includegraphics[width=0.56\textwidth]{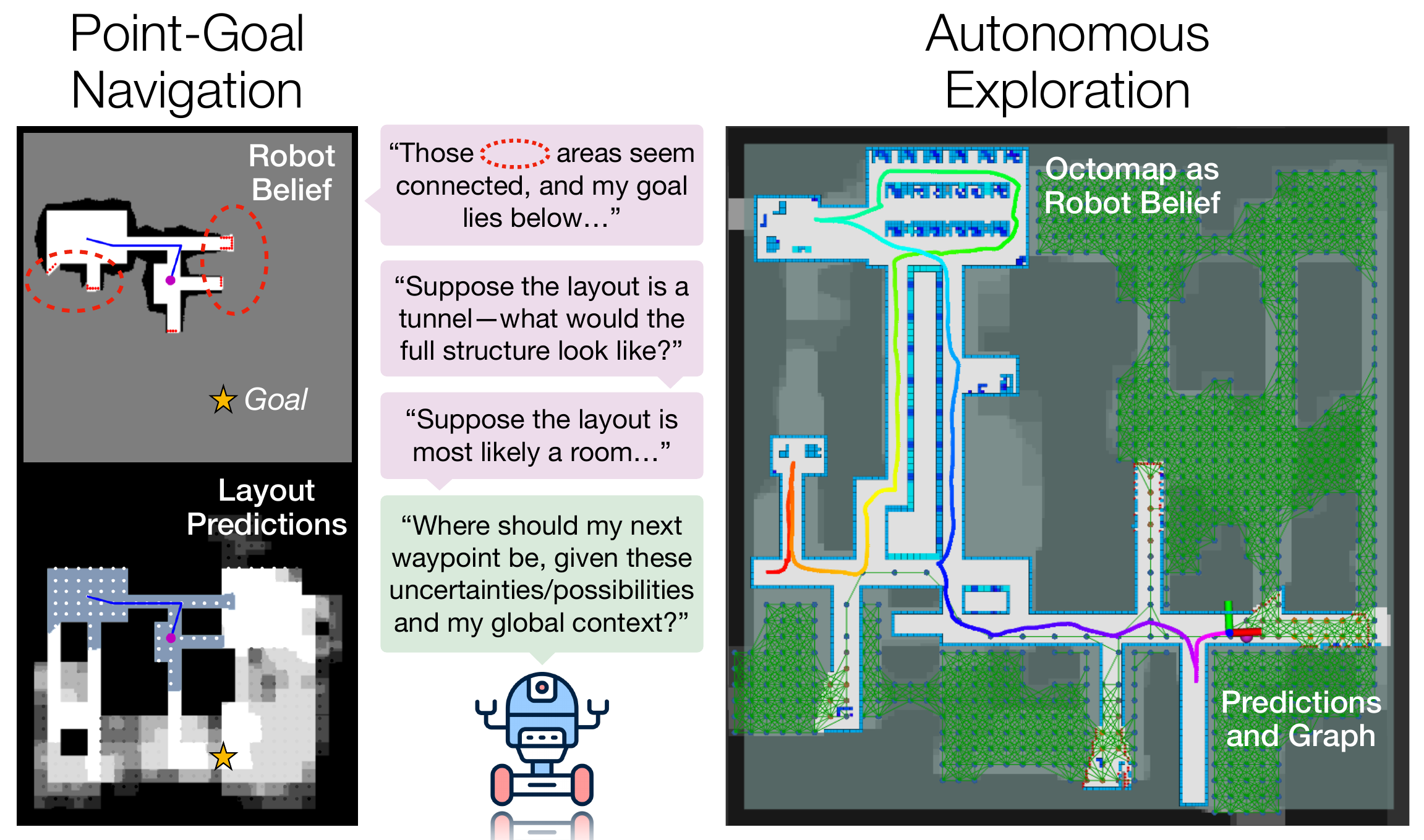}
    \caption{\textbf{CogniPlan's layout prediction and trajectory.}
    We show halfway navigation in a simulated map and halfway exploration in a Gazebo environment.
    }
    \vspace{-0.5cm}
    \label{fig:teasing}
\end{wrapfigure}

Path planning is a fundamental capability of robotic intelligence, enabling mobile robots to compute efficient routes toward designated targets to accomplish their missions.
In practical scenarios, however, environments are usually unknown beforehand, posing inherent challenges for decision-making.
This work focuses on two essential and coupled tasks: autonomous exploration and point-goal navigation, where the robot must plan in unknown environments to either fully map the space or reach a specific target, both via the shortest possible routes.

To reason effectively under such uncertainty in unknown environments, the robot must act on its partial and dynamically updating belief.
At the same time, it must estimate potential information gain, where seemingly ambiguous choices may lead to significantly different long-term outcomes.
However, most planning policies remain rule-based, typically selecting frontiers (i.e., the boundary between known traversable and unknown regions) using greedy heuristics based on proximity and/or expected unmapped volume~\citep{yamauchi1997frontier,gonzalez2002navigation,bircher2016receding,karaman2011sampling}.
While effective in simple scenarios, such approaches react passively to observations without actively reasoning about or predicting the broader environmental layout, which limits their ability to plan with non-myopic spatial coherence.
Data-driven approaches, on the other hand, include deep reinforcement learning (DRL) methods that implicitly encode spatial knowledge within policy networks~\citep{zhu2018deep,li2019deep,cao2024deep,tai2017virtual,liang2023context}, as well as those explicitly predicting layouts to then apply conventional planning~\citep{shrestha2019learned,luperto2021exploration,georgakis2022uncertainty,sharma2024pre,ho2024mapex}.
However, these two paradigms, implicit policy learning and explicit layout prediction, are rarely integrated, leaving their potential synergy underexplored.
This work addresses the question: Can layout predictions be \textit{explicitly} harnessed to guide planning under uncertainty, so that a learned planner can reason over multiple plausible spatial hypotheses?

When humans plan paths, their brain typically constructs a \textbf{cognitive map}: an internal spatial representation of the environment that supports reasoning, memory recall, and forward planning during navigation~\citep{epstein2017cognitive}.
Inspired by this capability, we introduce \textbf{CogniPlan}: a fully learned path planning framework guided by multiple plausible layout predictions generated through \underline{co}nditional \underline{g}e\underline{n}erative \underline{i}npainting.
We illustrate this cognitive reasoning process in Fig.~\ref{fig:teasing}, where our robot first explicitly predicts the layout structure and then implicitly reasons about and plans its path under uncertainty.

We first adopt a Wasserstein generative adversarial network (WGAN)~\citep{gulrajani2017improved} to generate plausible predictions conditioned on ground-truth layout types.
We tailor our WGAN to be lightweight, allowing efficient repeated inference for generating diverse predictions that capture layout uncertainty.
The output is further post-processed to produce binary outputs indicating either free space or obstacles.
During forward inference, we input a set of layout conditioning vectors to generate multiple plausible predictions, each corresponding to a different structural assumption about the partially observed environment.
Guided by these layout hypotheses, CogniPlan's graph attention network (inspired by~\citep{cao2024deep}) reasons about potential uncertainty and information gain, allowing the agent to make real-time, informed decisions about the next waypoint for exploration or navigation.

Recent studies highlight graph attention-based approaches as flexible and scalable solutions for path planning, capable of generalizing across environments of varying sizes while maintaining a global context~\citep{cao2022catnipp,cao2024deep}.
However, since these policies usually rely solely on graph representations, they often fails to capture subtle yet meaningful details in the underlying occupancy map.
For example, frontiers with distinct shapes are encoded similarly in the graph, despite conveying critical structural information. 
CogniPlan addresses this limitation by leveraging a generative model to predict plausible layouts, effectively capturing non-trivial spatial details and amplifying them at the graph level to guide informed planning decisions.
We demonstrate a strong synergy between the scalability and reasoning capacity of the graph attention network and the high-fidelity spatial features captured by the generative model.
Compared to using our layout generator with conventional planners~\citep{cao2021tare,cao2023representation,hart1968formal} and to state-of-the-art DRL planners~\citep{cao2024deep,liang2023context}, CogniPlan achieves $17.7\%$ and $7.0\%$ reduction in travel length for exploration, and a $3.9\%$ and $12.5\%$ reduction for navigation, respectively, across hundreds of simulated maps.
We further evaluate CogniPlan on a realistic floor plan dataset, exhibiting higher efficiency without any additional training or fine-tuning.
Finally, we deploy our framework in a high-fidelity Gazebo simulation (in Appendix~\ref{appendix:gazebo}), as well as on a mobile ground robot, demonstrating its high-quality path planning and real-world applicability.

\section{Related Works}

\textbf{Exploration Planner.}
Classical exploration relies on hand-crafted heuristics. A seminal example is frontier-based exploration~\citep{yamauchi1997frontier}, where robots navigate to proximal frontiers to gradually uncover unknown space. Later work improved efficiency by integrating information gain metrics like frontier size and expected unmapped volume~\citep{holz2010evaluating,kulich2011distance,lu2020optimal}.
However, basic frontier-based algorithms are inherently greedy, often causing inefficient backtracking and myopic behavior in complex environments~\citep{visser2008beyond}. 
Sampling-based methods, rooted in the next-best-view paradigm~\citep{connolly1985determination}, alleviate these issues by optimizing longer-term trajectories~\citep{bircher2016receding,zhu2021dsvp}.
Advanced planners such as TARE~\citep{cao2021tare,cao2023representation}, HPHS~\citep{long2024hphs}, and FUEL~\citep{zhou2021fuel} leverage hierarchical structures to balance global planning and local refinement.
Learning-based methods typically use convolutional networks to encode the robot’s environmental belief~\citep{zhu2018deep,li2019deep,kamalova2022occupancy,tao2024learning}, preserving local geometry but suffering from limited receptive fields, which often forces planners to downsample or crop large environments.
Recent work therefore looked to graph attention networks as backbones to further boost reasoning and scalability~\citep{cao2024deep,chen2020autonomous}.
However, graph-based representations remain constrained to node-level features, overlooking fine-grained, high-fidelity spatial details critical for effective planning.

\textbf{Navigation Planner.}
Navigation planners can be broadly categorized into three groups: search-based, sampling-based, and learning-based methods.
Search-based methods (e.g., A*~\cite{hart1968formal}, D* \cite{likhachev2005anytime}, and D*Lite \cite{koenig2002dstar}) rely on graph expansion with heuristic functions, while sampling-based methods (e.g., RRT \cite{kuffner2000rrt}, RRT* \cite{karaman2011sampling}, and BIT* \cite{gammell2015bitstar}) explore the continuous space through incremental random sampling.
Despite their success in fully known environments, both categories struggle in unknown/partially known environments due to limited adaptability and lack of foresight.
On the other hand, learning-based methods, especially DRL-based, offer the potential to reason over uncertainty and plan efficiently without hand-crafted rules~\citep{zhang2017successor,everett2018motion,liang2023context,liang2024hdplanner}.
However, most navigation planners, similar to exploration ones, lack explicit spatial prediction of unknown regions, which limits their goal-directed behavior in complex scenarios~\citep{wei2021occupancy}. In this work, we explicitly model and leverage such predictions to improve planning efficiency and long-horizon navigation under uncertainty.

\textbf{Planning with Map Prediction.}
Predicting unknown layouts from observed environmental beliefs has emerged as a promising approach for exploration~\citep{luperto2021exploration,shrestha2019learned,ho2024mapex,tao2024learning,ericson2024beyond,baek2025pipe}, navigation~\citep{wei2021occupancy,katyal2019uncertainty}, or both~\citep{georgakis2022uncertainty,ramakrishnan2020occupancy}.
The use of deep learning methods, particularly generative models~\citep{gulrajani2017improved,suvorov2022resolution}, enables the prediction of semantically coherent and structurally plausible layouts, going beyond merely optimizing for statistical similarity to ground truth. 
Most existing methods rely on a \textit{single} deterministic layout prediction without modeling uncertainty and decouple prediction from planning.
They first generate a layout prediction and then compute information gain (e.g., predicted sensor coverage) to directly guide either a conventional planner~\citep{luperto2021exploration,shrestha2019learned} or a learning-based one~\citep{ramakrishnan2020occupancy,tao2024learning}.
More recent exploration planners such as UPEN~\citep{georgakis2022uncertainty}, MapEx~\citep{ho2024mapex}, and PIPE~\citep{baek2025pipe}, as well as the navigation planner in~\cite{katyal2019uncertainty}, incorporate uncertainty estimates from learned predictors to support more informed decision-making by \textit{conventional} planners.
CogniPlan bridges prediction and planning by providing a fully learning-based, uncertainty-aware framework for both exploration and navigation.
It estimates global layout uncertainty using multiple conditioning priors, which capture nuanced layout cues and amplify those important to downstream planning.
While methods like MapEx rely on a few predictions without ensuring full exploration, CogniPlan guides planning with its predictions while maintaining coverage, making it more broadly applicable in real-world scenarios.

\section{Framework Design}

An overview of the CogniPlan framework is presented in Fig.~\ref{fig:diagram}, illustrating how the conditional generative inpainting network collaborates with the graph attention planner network.
In this section, we explain how the fine-grained belief map $\mathcal{M}$, which contains non-trivial structural details/cues, is amplified through the layout predictions $\hat{\mathcal{M}}$.
Information relevant to planning, including frontiers, connectivity, and uncertainty guidance, is extracted and encoded into a graph representation $G' = (V', E)$ for use in downstream planning.

\subsection{Problem Formulation}

We consider a bounded environment $\mathcal{E}$, partitioned into free space $\mathcal{E}_f$ and occupied space $\mathcal{E}_o$.
The robot holds a spatial belief over the environment, which is discretized into a 2D occupancy map $\mathcal{M}\subset \mathbb{Z}^2$ for layout prediction and planning.
Following a policy $\pi$, the robot navigates within the traversable free space, forming a collision-free trajectory $\psi:\{0,1,2,...\} \rightarrow \mathcal{E}_f$.
Here, $\psi_i(t)$ indicates the robot's waypoint location at timestep $t$.
Upon observation with its onboard sensor (a LiDAR in our case) along $\psi$, the unknown areas $\mathcal{M}_u$ are gradually revealed and classified into either free areas $\mathcal{M}_f$ or occupied area $\mathcal{M}_o$, such that $\mathcal{M}=\mathcal{M}_u\cup\mathcal{M}_f\cup\mathcal{M}_o$.
The exploration task is considered complete upon the closure of $\mathcal{M}_o$ with respect to $\mathcal{M}_f$, while the navigation task is complete when the robot reaches the destination $\Bar{p}\in\mathcal{E}_f$.
Both planning tasks are formulated to minimize the cost of completion, measured by either travel length $D(\psi)$ or makespan.

\subsection{Conditional Generative Inpainting}

Image inpainting is a well-established technique for predicting missing regions in an image based on the surrounding context.
Most existing methods~\citep{luperto2021exploration,shrestha2019learned,georgakis2022uncertainty,ramakrishnan2020occupancy} utilize encoder-decoder architectures, such as variational autoencoder (VAE)~\citep{kingma2013auto} or U-Net~\citep{ronneberger2015u}, to generate local patches within occluded areas, limiting their use to refine the map and estimate information gain.
We hypothesize that, with a diverse set of layout predictions, even if individually inaccurate, a planner network can gradually learn to balance exploration and exploitation.
This allows it to reason about both potential information gain and the environment's topology, enabling non-myopic path planning.

\textbf{Learning to Inpaint Layouts.}
To generate diverse predictions, we train a WGAN with gradient penalty on a dataset comprising three subsets of maps (namely room, tunnel, and outdoor), each paired with a ground-truth layout type conditioning vector encoded as one-hot during training.
This layout conditioning vector $z$ is then expanded to match the spatial dimensions of the input and concatenated with the robot’s occupancy map belief and a mask indicating the unknown regions to be predicted.
The generator outputs a layout prediction $\hat{\mathcal{M}}=\mathrm{Gen}(\mathcal{M}\mid z)$ that is composed of $\hat{\mathcal{M}}_f$ and $\hat{\mathcal{M}}_o$, and is trained with the following objective weighted by $\lambda$:
\begin{equation}
\label{eq:gen}
\mathcal{L}_\mathrm{Gen} = 
- \mathbb{E}\big[\mathrm{Dis}(\hat{\mathcal{M}})\big] 
+ \lambda_1 \big\| \hat{\mathcal{M}} - \mathcal{E} \big\|_1 
+ \lambda_2 \big\| (\hat{\mathcal{M}} - \mathcal{E}) \odot M_\mathrm{sd} \big\|_1 \big/ \textstyle\sum M_\mathrm{sd}
- \lambda_3\, \mathrm{F1}(\hat{\mathcal{M}}, \mathcal{E}),
\end{equation}
which includes one adversarial loss from the discriminator and three reconstruction terms, including an L1 loss over the entire prediction and a Dice (F1-score) loss for overlap accuracy.
However, overemphasizing reconstruction loss in all unknown areas can mislead the training process.
Intuitively, missing pixels near known regions are less ambiguous and more critical for ensuring robot traversability.
Following~\citep{yu2018generative}, we implement a spatially discounted mask $M_\mathrm{sd}$ and extend it to a more flexible formulation that accommodates irregular mask shapes (known areas in $\mathcal{M}$).
We assign higher weights to regions closer to known areas and apply a normalization term to account for varying mask sizes.
The weights decay exponentially with distance from known regions, controlled by a decay factor (we use $\gamma=0.95$).
See Appendix~\ref{appendix:framework} for more implementation details.

\textbf{Layout Uncertainty during Inference.}
To account for the unknown layout type during inference, we design a set of diverse conditioning vectors $\mathcal{Z}=\{z_0, z_1, ...\}$, where each vector represents a specific structural pattern or a mixture thereof. 
This set includes both one-hot and soft one-hot variants, allowing the generator to produce a diverse mixture of layout predictions.
We post-process the output to obtain binary predictions and apply morphological closing and opening operations to fill small holes and remove minor noises/artifacts. 
Since each pixel is classified as either free or occupied, the average of multiple predictions can fully characterize the uncertainty, as it follows a Bernoulli distribution.
The inpainting model captures and amplifies non-trivial spatial cues in the occupancy map that are often overlooked by the graph representation, such as incomplete corridor connections, newly revealed wall corners detected by LiDAR, and multi-branch frontier structures emerging ahead of corridors.
We then incorporate the estimated uncertainty into the graph representation used by our planner.
Notably, our lightweight generator has fewer than 0.35M parameters, enabling real-time multiple forward inferences for uncertainty estimation, even when running entirely on CPU.
Its efficiency also facilitates the use of the generated layout predictions for subsequent downstream planner model training without introducing much computational overhead, in contrast to the finetuned LaMa inpainting model~\citep{suvorov2022resolution} used in~\citep{ho2024mapex,baek2025pipe}, which has over 50M parameters.

\begin{figure}[t]
\centering
\includegraphics[width=\textwidth]{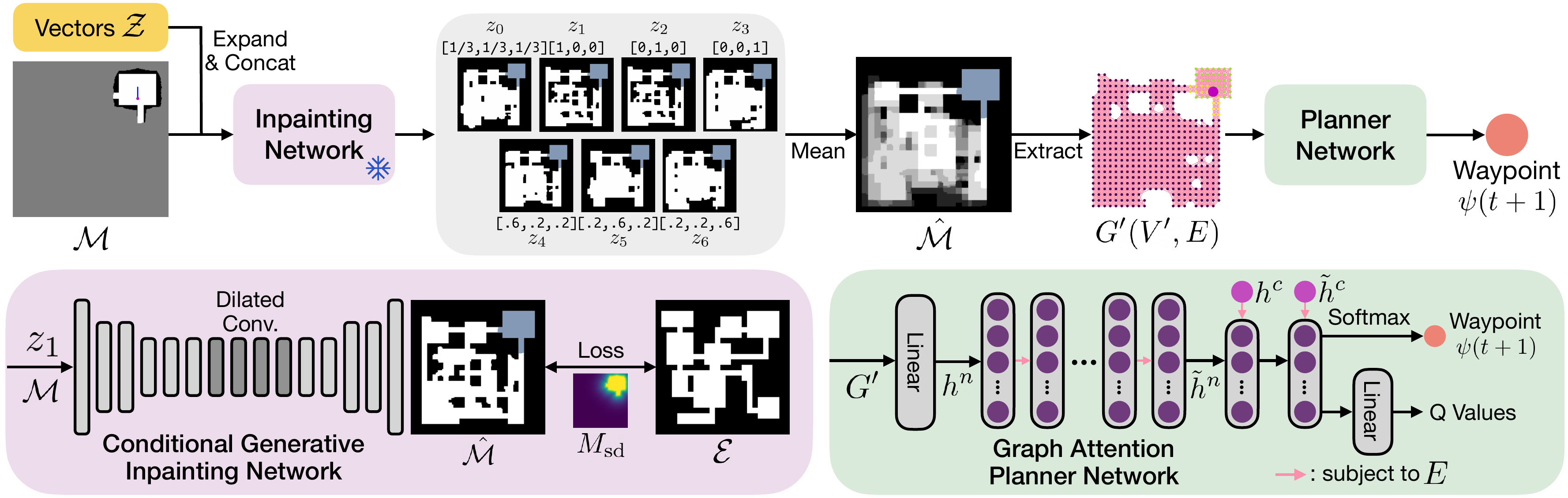}
\caption{
\textbf{CogniPlan framework.}
We first train a generative inpainting network on procedurally-generated maps, given their ground-truth layout type vector (room, tunnel, or outdoor), and then freeze the model to train a graph-attention-based planner network.
Our planner reasons over multiple predictions generated from a set of conditioning vectors by incorporating probabilistic information into the graph feature, and iteratively outputs the next waypoint for exploration or navigation.
}
\vspace{-0.4cm}
\label{fig:diagram}
\end{figure}

\subsection{Uncertainty-Guided Planner Network}

\textbf{Graph as Observation.}
Learning to plan on graphs offers a more scalable backbone than directly planning on occupancy map images, enabling better reasoning and reduced spatial complexity~\citep{cao2024deep,liang2023context}.
Instead of building a graph from the occupancy map belief $\mathcal{M}_t$, we construct a collision-free graph $G_t(V_t,E_t)$ at each decision timestep $t$, based on the layout predictions $\hat{\mathcal{M}_t}$.
We choose to uniformly distribute the nodes $v\in V$ over the free areas, and connect them based on proximity and line-of-sight visibility, forming edges $(v_i,v_j)\in E$.
In addition to the node coordinates, we associate each node with uncertainty and frontier-related attributes into the graph, yielding augmented node representations $v'_i=(v_i,s_i,p_i,u_i,g_i)\in V'$.
Specifically, 
\textbf{(1)} $s_i\in\{0,1\}$ is the signal indicating whether the node lies in known or predicted areas; 
\textbf{(2)} $p_i\in[0,1]$ is the average predicted probability of free areas. Note that the layout predictions are averaged to represent uncertainty; 
\textbf{(3)} $u_i\in[-1,1]$ is the normalized utility, based on the number of observable frontiers at the node position (with $u_i = -1$ assigned in predicted areas); 
and \textbf{(4)} $g_i\in\{0,1\}$ is the guidepost, indicating whether the node lies on the shortest trajectory to the nearest frontier within the known areas, which we find particularly useful for backtracking when the robot reaches a dead-end.
Note that we compute the guidepost on the predicted graph rather than solely within the known areas, which helps optimize the path toward the frontier.
Following~\citep{liang2023context}, for the navigation task, we additionally incorporate a direction vector toward the point-goal into the graph.
In this way, features from the predicted layout are encoded into an informative graph $G'_t(V'_t, E_t)$, which is then fed into the planner network.

\textbf{Policy Network.}
Our graph attention-based policy network comprises an encoder and a decoder: the encoder aggregates information from nodes across the known map, while the decoder selects the next neighboring node to visit based on the encoded global context.

The encoder consists of a feed-forward layer followed by $N$ stacked masked self-attention layers.
We first project the node properties in $V'_t$ into $d$-dimensional features $h^n$, which are then enhanced through attention layers that aggregate information from neighboring nodes, subject to the mask derived from graph connectivity $E_t$.
The mask constrains each node’s attentive field to immediate neighbors, while stacking multiple layers progressively expands the context to $N$-hop neighborhoods.
This enables the encoder to integrate both local structural information and broader spatial dependencies into a rich, uncertainty-aware representation $\tilde{h}^n$ of the graph/environment information.
The decoder determines the final policy based on $\tilde{h}^n$.
We first select the feature at the robot’s current position as the query, $h^c = \tilde{h}^{\psi(t)}$, and use it with neighboring node features (as key-value pairs) in an attention layer to produce an updated feature $\tilde{h}^c$.
A single-head attention mechanism is then applied, using $\tilde{h}^c$ and its neighboring node features, with the resulting attention scores directly serving as the output policy $\pi$~\cite{cao2022dan}, from which the next waypoint is selected.
It allows the policy to adapt to varying numbers of neighbors, accommodating arbitrary graph connectivity.

\textbf{Critic Network.}
The critic network estimates the state-action (Q) values based on the robot's partial observations, providing learning signals to guide the policy network.
We adopt privileged learning by directly feeding the ground-truth graph $G^*(V^*,E^*)$, extracted from $\mathcal{E}$, into the critic to improve value estimation, thereby facilitating more effective policy learning under a stochastic partially observable Markov decision process (POMDP).
The critic has an identical network architecture as the policy network, except that its final layer outputs Q-values instead of single-head attention scores.
Similarly, the critic contains prediction probability information $p_i$ as input, but only for nodes $v^*_i$ in the ground-truth graph.

\subsection{Network Training Details}

\textbf{Inpainting Model.}
Generative adversarial networks are prone to training instability, often resulting in issues such as mode collapse or gradient explosion.
To mitigate this, we adopt a WGAN with gradient penalty (WGAN-GP~\citep{gulrajani2017improved}) and apply a warmup phase during the initial 10k iterations, where only the generator is trained using a pure reconstruction loss (last three terms in Eq.~\ref{eq:gen}).
This warmup encourages the generator to produce coarse but meaningful layout estimates before engaging in full adversarial training for the subsequent 490k iterations.
The discriminator loss is given by:
\begin{equation}
\mathcal{L}_{\mathrm{Dis}}
= \mathbb{E}_{\hat{\mathcal{M}}}\bigl[\mathrm{Dis}(\hat{\mathcal{M}})\bigr]
- \mathbb{E}_{\mathcal{E}}\bigl[\mathrm{Dis}(\mathcal{E})\bigr]
+ \lambda_{\mathrm{gp}}\, \mathbb{E}_{\tilde{\mathcal{M}}}\left[\bigl(\|\nabla_{\tilde{\mathcal{M}}}\mathrm{Dis}(\tilde{\mathcal{M}})\|_{2}-1\bigr)^{2}\right],
\end{equation}
where $\tilde{\mathcal{M}}=\alpha \hat{\mathcal{M}}+(1-\alpha) \mathcal{E}$ is a random linear interpolation between layout prediction and the ground truth, with $\alpha \sim \mathcal{U}(0,1)$.
The gradient penalty term, weighted by $\lambda_\mathrm{gp}$, encourages the 1-Lipschitz constraint on the discriminator, following~\citep{gulrajani2017improved}. 
To build our training dataset for the inpainting network, we use TARE~\citep{cao2021tare,cao2023representation} to explore the Dungeon environment provided by~\citet{chen2019self}.
We procedurally generate 3000 random maps, evenly distributed across three layout types: room, tunnel, and outdoor environments.
During exploration, we monitor the exploration rate and save the occupancy map whenever the rate increases by more than $10\%$.

\textbf{Planner Model.}
Both the exploration and navigation planners are trained using the same inpainting model.
We train the exploration planner on another set of 6000 maps (2000 per environment type) and the navigation planner on 2000 maps with assigned start and target positions, using the soft actor critic (SAC) algorithm~\citep{haarnoja2018soft} with discrete actions.
Each map has a grid size of $250 \times 250$, with a sensing range of $40$ pixels.
Nodes are uniformly distributed at intervals of $9\mathrm{px}$ and connected to neighboring nodes within a radius of $26\mathrm{px}$ (connect to 25 neighbors at most in this setting).
Empirically, we use a set of seven layout conditioning vectors (i.e., $|\mathcal{Z}| = 7$), consisting of one uniform vector, three one-hot vectors, and three soft-biased vectors favoring each layout type.
Alternatively, our planner network can also be trained using a reduced set of four vectors or even a single uniform vector, although this leads to degraded performance.
Additional details are provided in Appendix~\ref{appendix:framework}.

\section{Experiments}

In this section, we evaluate CogniPlan’s exploration performance across two datasets and deploy it on real robot using the same trained model without any retraining or fine-tuning.
We also evaluate its navigation capability on simulated maps and compare both results against other baseline algorithms.

\subsection{Evaluation on Simulated Maps}

We evaluate CogniPlan on 150 maps for the exploration task, covering three different layout types in simulated environments, and on 100 maps for the navigation task.
All evaluation maps are unseen during training.
We measure and compare the travel length required to fully explore the environment or reach a designated point-goal.
We compare CogniPlan against baseline exploration planners:
\textbf{(1)} nearest: navigate to the nearest frontier; 
\textbf{(2)} utility: navigate to a frontier balancing proximity and utility;
\textbf{(3)} NBVP~\citep{bircher2016receding}: a sampling-based exploration planner;
\textbf{(4)} TARE Local~\citep{cao2021tare,cao2023representation}: the local sampling-based planner from TARE;
\textbf{(5)} ARiADNE+~\citep{cao2024deep}: an ablative improved ARiADNE variant with privileged learning and the same guidepost as CogniPlan.
\textbf{(6)} Inpaint+TARE: an ablative baseline applying the TARE coverage planner to inpainted predictions treated as ground truth.
For navigation, we compare against:
\textbf{(1)} BIT~\citep{gammell2015bitstar}: a sampling-based navigation planner; 
\textbf{(2)} D*Lite~\citep{koenig2002dstar}: a search-based navigation planner;
\textbf{(3)} CA~\citep{liang2023context}: an ablative context-aware DRL navigation planner;
\textbf{(4)} Inpaint+A*: an ablative method that uses our layout prediction and plans A* path to the closest unknown (if any) region to the goal.
For all results, we use CogniPlan with four layout predictions (i.e., $|\mathcal{Z}| = 4$), from exploration and navigation planner model trained with seven, to balance performance and computational efficiency.

We present the exploration results in Table~\ref{tab:expperf} and the navigation results in Table~\ref{tab:navperf}, where CogniPlan consistently outperforms all baselines, including state-of-the-art learning-based approaches, and slightly surpasses the near-optimal navigation planner D*Lite.
Compared to using inpainting with a conventional planner or learning-based methods without layout predictions (e.g., ARiADNE+ and CA), CogniPlan achieves a travel length reduction of $17.7\%$ and $7.0\%$ for exploration, and $3.9\%$ and $12.5\%$ for navigation.
Interestingly, Inpaint+TARE performs particularly poorly.
Upon closer inspection, we believe this may stem from high variability in layout predictions during the early stages of exploration, which causes the coverage planner to generate inconsistent and divergent trajectories.
The planner struggles to effectively handle uncertainty and becomes overly reliant on potentially inaccurate predictions, often leading to zig-zagging paths and reduced efficiency.
We believe this highlights the importance of using an uncertainty-aware downstream planner that can leverage layout uncertainty to guide more consistent and efficient planning.

\begin{figure}[t]
    \centering
    \begin{minipage}[t]{0.312\textwidth}
        \centering
        \includegraphics[width=\textwidth]{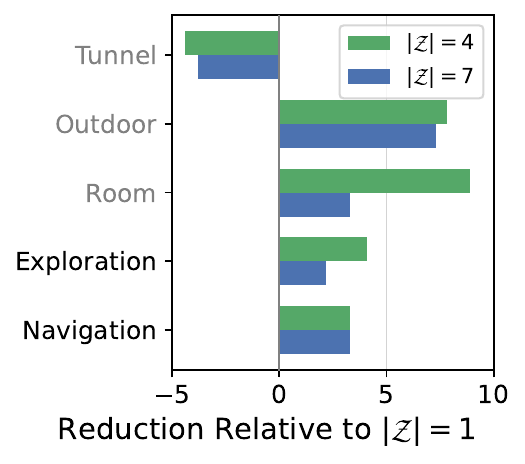}
        \vspace{-0.65cm}
        \caption{\textbf{Travel length reduction.}
        Comparison of 4 and 7 predictions vs. 1.}
        \label{fig:npred}
    \end{minipage}
    \hfill
    \begin{minipage}[t]{0.322\textwidth}
        \centering
        \includegraphics[width=\textwidth]{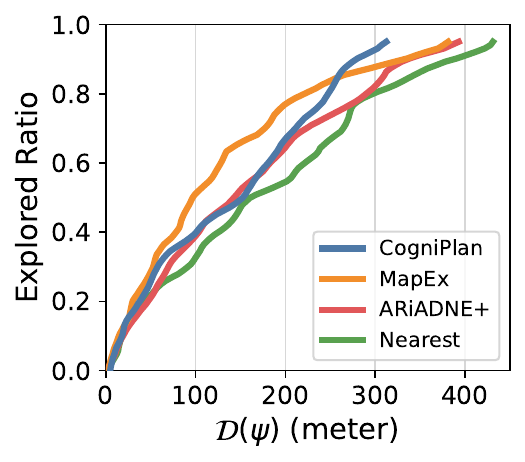}
        \vspace{-0.7cm}
        \caption{\textbf{Exploration progress in floor plans.}
        Average exploration rate vs. travel distance.}
        \label{fig:curve}
    \end{minipage}
    \hfill
    \begin{minipage}[t]{0.322\textwidth}
        \centering
        \includegraphics[width=\textwidth]{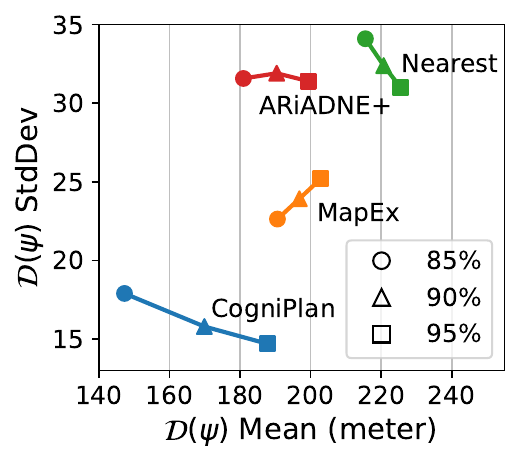}
        \vspace{-0.7cm}
        \caption{\textbf{Robustness to random starts.} Travel length at different exploration rates.}
        \label{fig:robustness}
    \end{minipage}
    \vspace{-0.5cm}
\end{figure}

To investigate the impact of the number of predictions, we evaluate CogniPlan using $|\mathcal{Z}| = 1$, $4$, $7$ predictions for both exploration and navigation tasks.
As shown in Fig.~\ref{fig:npred}, we report the reduction in travel length when using four or seven predictions compared to a single prediction.
The layout types apply only to exploration.
We observe that increasing the number of predictions during inference generally improves performance.
However, the gains tend to saturate as the number of predictions increases, suggesting that a small number of diverse hypotheses is sufficient to guide planning.

\begin{table}[t]
\centering
\caption{
\textbf{Exploration performance comparison over 150 maps (50 per environment).}
We report the mean and standard deviation of travel length to complete exploration (lower is better).
}
\label{tab:expperf}
\fontsize{8.5}{10}\selectfont
\setlength{\tabcolsep}{2pt}
\begin{tabular}{c|c|c|c|c|c|c|c}
\toprule
& Nearest & Utility & NBVP & TARE Local & ARiADNE+ & Inpaint+TARE & CogniPlan \\
\midrule
Room     
& $412.2(\pm 51.3)$ & $397.1(\pm 63.3)$ & $409.5(\pm 63.4)$ & $400.2(\pm 55.3)$ & $379.5(\pm 54.5)$ & $436.5(\pm 109.0)$ & $\textbf{360.1}(\pm 43.5)$ \\
Tunnel   
& $297.8(\pm 69.3)$ & $293.7(\pm 73.6)$ & $303.5(\pm 87.1)$ & $300.4(\pm 75.2)$ & $278.4(\pm 66.7)$ & $319.7(\pm 85.3)$ & $\textbf{268.5}(\pm 58.9)$ \\
Outdoor  
& $372.7(\pm 67.6)$ & $358.6(\pm 60.3)$ & $318.8(\pm 72.9)$ & $310.9(\pm 56.0)$ & $330.1(\pm 52.9)$ & $360.9(\pm 73.0)$ & $\textbf{290.7}(\pm 50.8)$ \\
\midrule
Average  
& $360.9$           & $349.8$           & $343.9$           & $337.2$           & $329.3$           & $372.3$           & $\textbf{306.4}$ \\
\hline
Gap & $15.1\%$ & $12.4\%$ & $10.9\%$ & $9.1\%$ & $7.0\%$ & $17.7\%$ & $0\%$ \\
\bottomrule
\end{tabular}
\vspace{-0.5cm}
\end{table}

\begin{table}[t]
\centering
\caption{
\textbf{Navigation performance comparison over 100 maps.}
We report the mean and standard deviation of travel length to reach the point goal (lower is better).
}
\label{tab:navperf}
\fontsize{8.5}{10}\selectfont
\setlength{\tabcolsep}{5pt}
\begin{tabular}{c|c|c|c|c|c}
\toprule
& BIT & D*Lite & CA & Inpaint+A* & CogniPlan \\
\midrule
Distance   
& $275.2 (\pm 130.4)$ & $224.6(\pm 85.9)$ & $253.9 (\pm 112.4)$ & $231.1 (\pm 125.8)$ & $\textbf{222.1} (\pm 64.7)$ \\
\hline
Gap & $19.3\%$ & $1.1\%$ & $12.5\%$ & $3.9\%$ & $0\%$ \\
\bottomrule
\end{tabular}
\vspace{-0.5cm}
\end{table}

\subsection{Evaluation on Realistic Floor Plans}

To benchmark CogniPlan in a more realistic setting, we further evaluate it on the KTH floor plan dataset~\citep{aydemir2012can} and on Gazebo environments (Appendix~\ref{appendix:gazebo}), which, unlike standardized simulated maps suited for large-scale training and testing, better reflects real-world indoor environments.
We compare CogniPlan against the following representative baseline methods in a zero-shot setting:
\textbf{(1)} nearest;
\textbf{(2)} ARiADNE+;
\textbf{(3)} MapEx~\citep{ho2024mapex}, a non-learning approach that uses map predictions and is trained on the KTH dataset to estimate information gain, representing the state-of-the-art for fast exploration on these maps.
For fairness, we tune the parameters of all methods to maximize their performance on those maps and ensure consistency with those used in CogniPlan (where applicable).

We first deploy each method across 10 diverse floor plans, running it twice on each map with different starting positions.
We record the exploration rate and travel length $D(\psi)$, and report the average exploration progress curves in Fig.~\ref{fig:curve}.
We observe that MapEx achieves the best performance during the early stages of exploration (up to $80\%$), but its progress plateaus when approaching full coverage. 
This is likely because MapEx prioritizes waypoints that are likely to yield high information gain; however, its effectiveness diminishes in the final stages, where the remaining unexplored regions tend to be smaller, more isolated, or harder to predict.
Although these out-of-distribution layouts differ significantly from the data used to train CogniPlan's inpainting and planner models, leading to some predictions that visibly deviate from the ground truth, CogniPlan still maintains a steady exploration, outperforming other methods beyond $90\%$ coverage.
We believe this may be because the predicted layouts, though coarse, provide informative structural priors that enable long-horizon planning and reduce redundant local actions, which in turn enables the near-linear exploration progress shown in Fig.~\ref{fig:curve}.
This is further supported by CogniPlan outperforming ARiADNE+, which does not explicitly predict environment layouts.

CogniPlan also demonstrates robustness to starting position variability in exploration. 
We evaluate performance on two similarly sized maps, each with five diverse starting points selected from different spatial regions. 
We report the mean and standard deviation of total travel distance at $85\%$, $90\%$, and $95\%$ exploration coverage, which reflect near-complete exploration performance.
As shown in Fig.~\ref{fig:robustness}, CogniPlan achieves the lowest average travel distance and standard deviation, indicating reduced sensitivity to starting position and more globally consistent planning (the closer to the bottom left corner, the better). 
At $90\%$ and $95\%$ coverage, it reduces standard deviation of $D(\psi)$ by over $50\%$ compared to nearest and ARiADNE+.
We attribute this improvement to the predictions providing global structural priors that guide the planner, reducing backtracking and redundant revisits.

\subsection{Evaluation on Hardware}

\begin{wrapfigure}{r}{0.55\textwidth}
    \centering
    \vspace{-1.2cm}
    \includegraphics[width=0.55\textwidth]{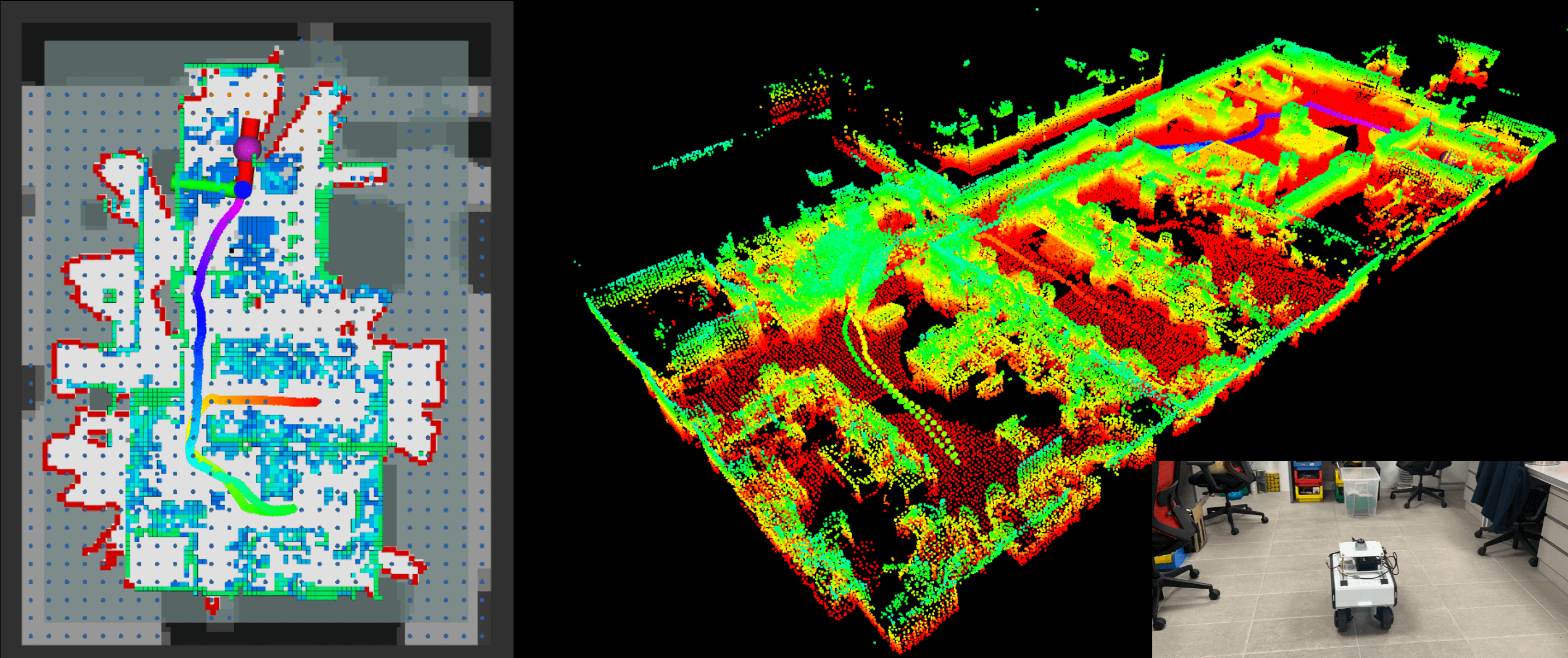}
    \caption{\textbf{Real-world exploration experiment in a $\textrm{30m} \times \textrm{10m}$ indoor laboratory.}
    We show our mobile robot, the lab environment, the intermediate Octomap, and the final point cloud with the robot's trajectory.
    }
    \vspace{-0.3cm}
    \label{fig:realrobot}
\end{wrapfigure}

We validate CogniPlan through real-world exploration experiments conducted in a $30\text{m} \times 10\text{m}$ indoor laboratory filled with cluttered objects such as tables, chairs, equipment, and moving people. 
Our mobile platform is a four-wheeled omnidirectional robot with a maximum speed set to $0.75\,\text{m/s}$, equipped with a Livox Mid-360 3D LiDAR for odometry and mapping.
We use an Octomap resolution of $0.2\text{m}$ per cell and a waypoint node resolution of $0.8\text{m}$.
As observed, the robot generally completes exploration within $6$ minutes, with a travel length of approximately $100$ meters. 
Fig.~\ref{fig:realrobot} illustrates how the robot predicts layouts during exploration.
We believe these results show the promise of CogniPlan's framework for practical use.

\section{Conclusion}

In this work, we proposed CogniPlan, a path planning framework that combines uncertainty from multiple explicit layout predictions with a graph attention-based planner trained via deep reinforcement learning. 
We demonstrate their synergistic effect, where the inpainting model amplifies non-trivial map details into uncertainty-aware graph representations to guide downstream planning.
We show CogniPlan plan more efficient paths for both exploration and navigation tasks across simulated environments and realistic floor plans, all without additional training.
We further deploy it on a mobile robot, showcasing its ability to handle real-world scenarios.

\clearpage

\section*{Limitations and Future Work}

\textbf{Generalization and Prediction Diversity.}
Although CogniPlan's planner network leverages graph attention to scale across arbitrary graph topologies, the inpainting model still requires a fixed-size occupancy map input.
This imposes certain prior assumptions on the environment, which may limit the scalability and generalization of the overall framework, and leave a domain gap upon deployment.
A possible solution is to adopt resolution-agnostic network architectures, coupled with more diverse training data, to enhance spatial understanding across varied environments.

\textbf{One-Way Pipeline.}
After pre-training our inpainting model, the generator is frozen during planner training.
That is, the planner must adapt to whatever predictions it receives, which limits continual improvement and hampers adaptation to novel layouts discovered online.
To address this, we plan to explore end-to-end joint training, allowing planner gradients to flow into the generator, thereby improving both components simultaneously, although this approach requires careful architectural design.
We also plan to use active sampling to select conditioning vectors in latent space, instead of relying on a fixed or handcrafted set.

\textbf{Replanning Computation Overhead.}
We employ a lightweight inpainting model capable of rapidly inferring the layout; however, the main bottleneck lies in the graph rescheduling, where each new prediction triggers a complete rebuild of the robot's belief graph.
This process can be accelerated by comparing successive predictions and incrementally updating the graph through selective addition and pruning.

\appendix

\section{Validation in Gazebo Simulation}
\label{appendix:gazebo}

\textbf{Environment Setup}
We validate CogniPlan's real-world performance in a high-fidelity Gazebo environment provided by~\citet{cao2021tare,cao2023representation}, using a four-wheel-drive mobile robot equipped with a 16-line Velodyne LiDAR for odometry and mapping.
The environment incorporates realistic motion constraints, while a local planner ensures safe execution of waypoints generated by the high-level planner. 
During validation, we adopt the graph rarefaction strategy from~\citep{cao2024deep} to reduce graph complexity. 
Specifically, we preserve a dense graph structure in predicted regions while applying rarefaction in known areas, and remove nodes that are not connected via a valid path to the robot's current node.

\begin{figure}[h]
    \centering
    \includegraphics[width=\linewidth]{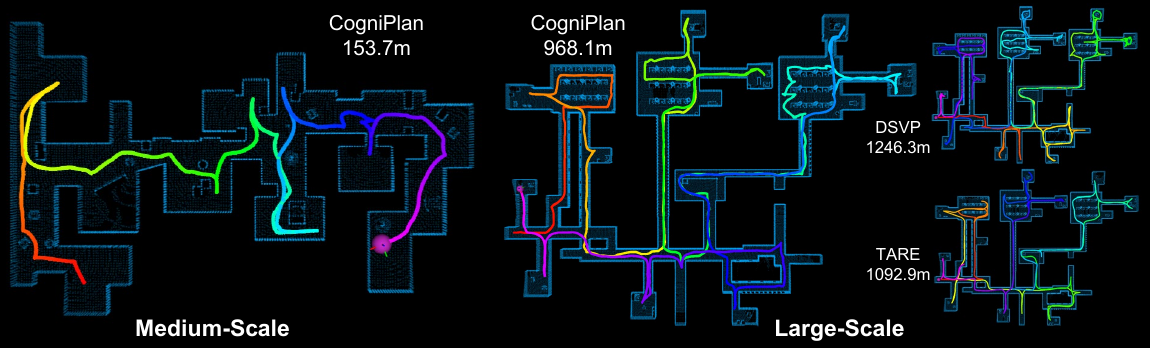}
    \caption{\textbf{Trajectories of CogniPlan and baseline planners in medium- and large-scale environment.}
    Colored lines represent the robot’s motion trajectories, with the red-to-purple spectrum indicating progression from start to end.
    }
    \label{fig:gazebo-traj}
\end{figure}

\textbf{Medium-Scale Environment.}
We first evaluate CogniPlan in a $54\rm{m}\times34\rm{m}$ indoor environment crafted by~\citet{long2024hphs}, which contains dense clutter and obstacles.
They also introduced HPHS, a hierarchical planner that directly samples from LiDAR data.
We compare CogniPlan against the baselines tested and reported in~\citep{long2024hphs}, including a frontier-based approach~\citep{orvsulic2019efficient} and an improved RRT planner TDLE~\citep{zhao2023tdle}.
The results are shown in Table~\ref{tab:medcomp}, where we observe that CogniPlan produces shorter paths with lower variance, indicating more consistent and stable trajectories due to effective layout/uncertainty prediction.
The corresponding trajectory is visualized in Fig.~\ref{fig:gazebo-traj}.

\begin{table}[h]
\small
    \centering
    \caption{\textbf{Comparison over 10 runs in the medium-scale environment.} We report the mean and standard deviation of travel distance and path efficiency relative to the explored area.}
    \begin{tabular}{c|c|c|c|c|c}
    \toprule
       Method  & Frontier-based & TDLE & TARE & HPHS & CogniPlan \\
    \midrule
      $D(\psi)$ $($\rm{m}$)$ & $270.6 (\pm 42.1)$& $290.5(\pm 35.6)$& $190.3(\pm 31.4)$& $176.9(\pm 24.7)$& $\textbf{170.9}(\pm 18.1)$\\
      Efficiency ($\rm{m^2/m}$) & $3.14$& $2.93$& $4.31$& $4.78$& $\textbf{4.95}$\\
    \bottomrule
    \end{tabular}
    \label{tab:medcomp}
\end{table}

\begin{figure}[h]
    \centering
    \begin{minipage}[b]{0.33\textwidth}
        \centering
        \includegraphics[width=\textwidth]{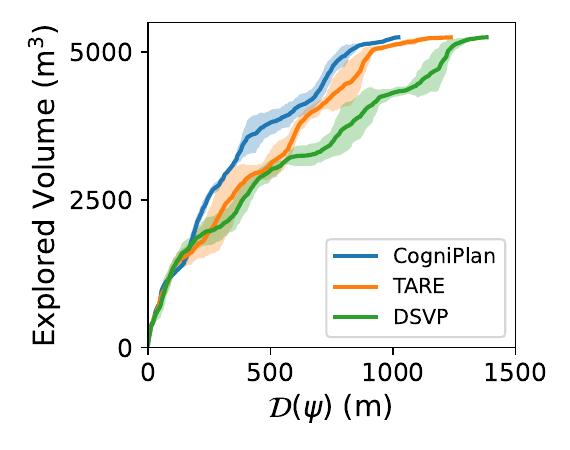}
        \caption{\textbf{Exploration progress in a large-scale environment.}}
        \label{fig:gazebo-curve}
    \end{minipage}
    \hfill
    \begin{minipage}[b]{0.65\textwidth}
        \centering
        \includegraphics[width=\textwidth]{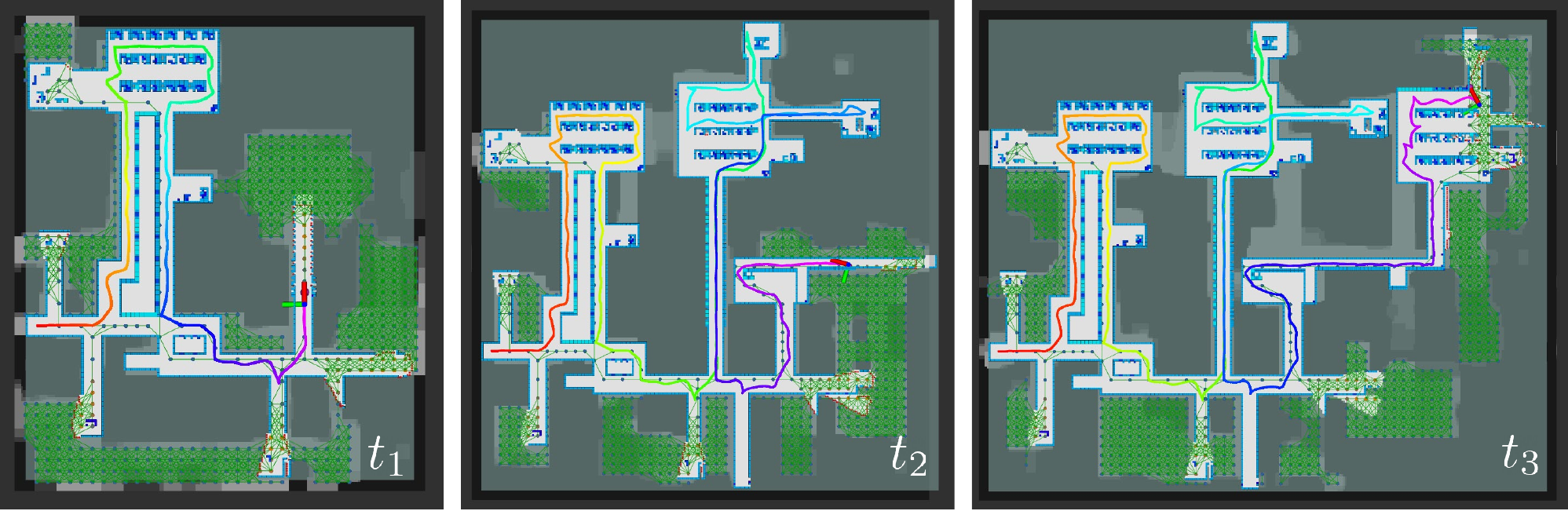}
        \caption{\textbf{Layout predictions at three different timesteps during exploration in a large-scale environment.}
        We also show the corresponding occupancy map (Octomap) and the rarefied graph (nodes and green edges).}
        \label{fig:gazebo-pred}
    \end{minipage}
\end{figure}

\textbf{Large-Scale Environment.}
We further evaluate our method in a $130\rm{m}\times100\rm{m}$ indoor setting comprising long, narrow corridors interconnected with spacious lobby areas. 
We compare CogniPlan against DSV Planner~\cite{zhu2021dsvp} and TARE~\cite{cao2021tare}, both of which represent state-of-the-art conventional planner for large-scale environments.
We run each method five times and present the exploration progress in Fig.~\ref{fig:gazebo-curve} and trajectory example in Fig.~\ref{fig:gazebo-traj}.
We notice that CogniPlan consistently outperforms TARE and DSVP throughout the exploration process, with notably lower variance, highlighting that layout predictions can enhance both performance and robustness.
Figure~\ref{fig:gazebo-pred} visualizes the layout predictions during exploration. 
Despite the environment being significantly out-of-distribution from the training data of both the inpainting and planner models, the predicted layouts still effectively support planning and contribute to improved performance.

We believe this demonstrates the potential of the CogniPlan framework as a fully learning-based planning approach that enhances performance, path robustness, and explainability. 
It also opens up promising directions for future research in multi-agent planning and visual navigation.

\begin{table}[h]
\small
    \centering
    \label{tab:largecomp}
    \caption{\textbf{Comparison over 5 runs in the large-scale environment.} We report the mean and standard deviation of travel distance and path efficiency relative to the explored volume.}
    \begin{tabular}{c|c|c|c}
    \toprule
       Method  & DSVP & TARE & CogniPlan \\
    \midrule
      $D(\psi)$ $($\rm{m}$)$  & $1300.6(\pm 58.7)$ & $1144.9(\pm 70.7)$ & $\textbf{999.1}(\pm 17.1)$ \\
      Efficiency ($\rm{m^3/m}$) & $4.03(\pm 0.19)$ & $4.58(\pm 0.30)$ & $\textbf{5.25}(\pm 0.09)$ \\
    \bottomrule
    \end{tabular}
\end{table}

\section{Framework Design Choices}
\label{appendix:framework}

\begin{figure}[h]
    \centering
    \includegraphics[width=0.7\linewidth]{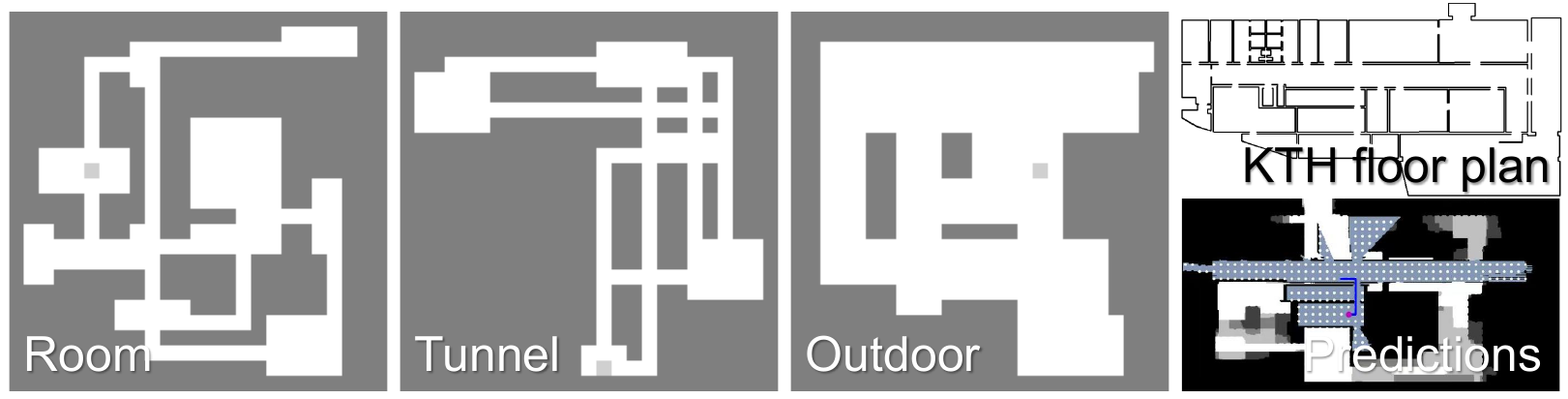}
    \caption{Examples of each environment type, along with an out-of-distribution prediction on the KTH floor plan.}
    \label{fig:layout}
\end{figure}

\textbf{Environment Types.}
\textit{Room} consists of spaces with varying sizes and connectivity. 
\textit{Tunnel} comprises long corridors, where the effects of each planning decision are significantly delayed.
\textit{Outdoor} is spacious with few obstacles or walls (see Fig.~\ref{fig:layout}).
We identify these three layout types as distinct and use them to train our inpainting network accordingly. 
Thus, we preserve these characteristics during planner evaluation as well. 

\textbf{Reconstruction Loss.} 
The reconstruction term comprises L1 (pixel‑level), Dice (region‑level), and an additional L1 loss that is spatially discounted around the known areas. 
The L1 and Dice losses are complementary: L1 alone stabilizes gradients but often causes artifacts, whereas adding a small amount of Dice loss reduces these artifacts and promotes spatial connectivity in the predictions.

\textbf{Conditioning Vector.}
We explored various conditioning strategies, including VAE sampling and Monte Carlo dropout for probabilistic approximation, and found that one-hot and soft one-hot vectors yield the best diversity and plausibility.
Conditioning vectors that are not valid probability distributions tend to produce unrealistic layouts (nearly all free space or all obstacles).
The vector choices for $|\mathcal{Z}|=7$ are illustrated in Fig.~\ref{fig:diagram}.
We uniformly average predictions without weighting (Eq.~\ref{eq:average}), as diversity naturally decreases as the robot explores/observes more of the environment. 
That is, early predictions primarily rely on layout conditioning vectors, while later ones converge based on the robot’s belief, forming a sense of conditional posterior.
\begin{equation}
\label{eq:average}
    \hat{\mathcal{M}}_\text{all}=\frac{1}{|\mathcal{Z}|} \sum_{z_i \in \mathcal{Z}} \operatorname{Gen}\left(\mathcal{M} \mid z_i\right)
\end{equation}

\textbf{Reward functions.}
We address both exploration and navigation by leveraging conditional layout predictions to guide the graph attention network backbone.
The two tasks share the same planner framework and both aim to complete the task with minimal travel length or makespan, but they differ in their input node features and in parts of the reward functions used to train the planner.
Following~\citep{cao2024deep} for exploration and~\citep{liang2023context} for navigation, both tasks reward task completion and penalize travel distance.
Exploration further rewards the discovery of new frontiers, while navigation rewards proximity to the goal, providing denser feedback that facilitates deep RL training.

\bibliography{ref}

\end{document}